# Empirical Performance Analysis of Conventional Deep Learning Models for Recognition of Objects in 2-D Images

*Sangeeta Satish Rao[a], Nikunj Phutela[b], V R Badri Prasad[c]\**

[a]*Student, PES University, 100 Feet Ring Road, Banashankari Stage III, Dwaraka Nagar, Banashankari, Bengaluru, Karnataka 560085, India*
[b]*Student, PES University, 100 Feet Ring Road, Banashankari Stage III, Dwaraka Nagar, Banashankari, Bengaluru, Karnataka 560085, India*
[c]*Associate Professor, PES University, 100 Feet Ring Road, Banashankari Stage III, Dwaraka Nagar, Banashankari, Bengaluru, Karnataka 560085, India*

A B S T R A C T

Artificial Neural Networks, an essential part of Deep Learning, are derived from the structure and functionality of the human brain. It has a broad range of applications ranging from medical analysis to automated driving. Over the past few years, deep learning techniques have improved drastically - models can now be customized to a much greater extent by varying the network architecture, network parameters, among others. We have varied parameters like learning rate, filter size, the number of hidden layers, stride size and the activation function among others to analyze the performance of the model and thus produce a model with the highest performance. The model classifies images into 3 categories, namely, cars, faces and aeroplanes.

## 1. Introduction

Object detection is an elementary Computer Vision task which deals with the classification of objects in a digital image to a particular class (such as airplanes, cars, humans, etc). This can further be used in the implementation of real-world systems like face detection, pedestrian detection, automated driving systems, video surveillance, among other applications. It has gathered a lot of attention in the last few years since it is closely related to video analysis. They also help provide keen insights on the image contents. In recent years, deep learning methods have gained momentum and are now able to learn large amount of features, at comparatively deeper levels, and are thus able to address the problems faced earlier in traditional network architectures, such as artificial neural networks. The most common method used for object detection in deep learning is Convolutional Neural Networks.

Convolutional Neural Networks are used to solve complex problems which were not possible by conventional neural networks, such as artificial neural networks. Now, with the help of Convolutional Neural Networks, tasks such as image classification, image recognition, and object detection, among others, can easily be performed. Convolutional Neural Networks are preferred to Artificial Neural Networks because of automatic feature extraction. Each variation in the parameters of the neural network results in large variations in the model, which may ultimately produce a net positive or net negative changes in the model's performance.

Convolutional Neural Networks take an input image, process it and classify it under certain categories. It is a deep learning algorithm that takes in an input image, assigns weights and biases to the various aspects in the image, and based on these parameters; the objects are differentiated from each other. The architecture of a Convolutional Neural Network is defined in a manner that is analogous to the organization of neurons in the human brain. The main role of a convolutional neural network is to reduce the input images to a form that is easier to process, without losing features that are vital to producing a good prediction of the object.

In this paper, we have varied parameters like learning rate, filter size, the number of hidden layers, stride size and the activation function among others to analyze the performance of the model and thus produce a model with the highest performance. In this paper, we have assumed the key metric indicative of the performance is the accuracy of the model. Our model was run on a highly optimized GPU to help boost the performance of the model, while also reducing the training time of the model. In addition to the conventional train-test split model, we also used cross-validation techniques like K-Fold and Leave One Out Cross Validation to analyze the performance of the model. We noticed that using such techniques produced a significant improvement in the performance of the model. Our model classifies the input image into one of the three categories, cars, faces and aeroplanes. The multiple class classification model has an accuracy of 68% while using the train-test split method, which is the conventional method. To increase the accuracy, as mentioned earlier, we used cross-validation techniques such as K-fold and Leave One Out Cross Validation techniques which produced accuracies of

\* *Sangeeta Satish Rao* Tel.:+919986170011
E-mail address: sangeetasatishrao@pesu.pes.edu

99.3% and 98.6 % respectively. In this paper, we have considered accuracy as the factor which suggests the performance of the model-as the loss of the model decreases, the accuracy of the model increases.

## 2. Literature Survey

### 2.1 Neural Networks

Manish M., Monika S.[15] illustrates what neural networks are and how ubiquitous they are in recent times. The paper covers the basics of Neural Networks, which gives a good foundation for beginners. Simple neural networks, however, are not usually used for Object Recognition as Convolutional Neural Networks yield better results for the task at hand.

S. Albawi, T. A. Mohammed and S. Al-Zawi[1] describe the basics of a convolutional network, which begins with an understanding of the basic elements such as convolution, stride size, padding and how features are extracted from Convolutional Neural Networks. The paper also explains the use of the non-linearity and pooling layers. It shows us how the final output is produced from the neural network from the fully connected layer. This paper also gives keen insights into the factors that affect the performance of a CNN, while also illustrating the common neural network architectures used in practice and their primary uses.

### 2.2 Object Recognition

Sudarshan D.[4] describes the convolutional network used to classify the images present in the CIFAR-10 dataset using Convolutional Neural Network. The authors used TensorFlow CPU to classify 60000 images using a Convolutional Neural Network. The dataset also contains images created by data augmentation, primarily flipping and changing the brightness in the image. The model is then run with the newly created images and the pre-existing images as the training dataset, and the testing set held separately, which is not subject to data augmentation. At the end of 25 epochs, the training accuracy is 96%, when a single image is passed to the model for testing.

Alex Krivhesky, Ilya Sutskever and Geoffrey E. Hinton[12] perform the task of image classification on the famous ImageNet dataset using a highly optimised GPU. The model used in this paper achieved an award-winning top-5 test error rate of 15.3%. The less error rate can be attributed to dropout being used in all layers in the feed-forward convolutional network. The model used in this paper employs a Convolutional Neural Network since their depth and breadth can be varied to a great extent, and they help make strong and almost always produce right predictions for images.

## 3. Methodology

### 3.1 Flow of Convolutional Neural Network

The convolution layer is the first layer which extracts data from the input image. The convolution helps preserve the relationship between pixels by learning image features using small squares of input data. The convolutional layer is a mathematical function which takes two inputs, namely the image matrix, and a filter or kernel. The stride size defines the number of pixels the filter moves at a time-if the stride size is 1, the filter moves by 1, and so on. The greater the stride size, the smaller the output produced from the filter. Once the output of the filter is obtained, it is passed to the max-pooling layer, which is typically employed to reduce the number of parameters when the images are too large. The output of the max-pooling layer is then fed to the fully connected layer, which helps flatten the matrix into a vector. The features obtained from the fully connected layer, are combined to create a model which finally classifies the images into 3 categories, namely cars, faces, and aeroplanes.

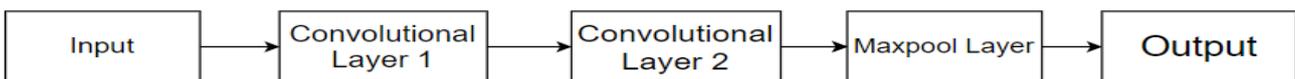

Fig. 1 –Flow of Convolutional Neural Network.

*3.2 Datasets*

The model used in this paper classifies objects into 3 categories, namely, cars, faces, and airplanes.
- Cars: The dataset used for the cars category is the Stanford Cars dataset. The Stanford Cars dataset consists of 16185 images, out of which 8144 images for the training phase, and 8044 images were selected for the testing phase. The various images in this dataset are from different makes and models, such as the 2012 Tesla Model S and the 2012 BMW M3 coupe, among others.
- Faces: The dataset used for the faces category is the Caltech Faces dataset. The Caltech faces dataset consists of 10524 images, which have been collected from the web using Google Image search. The information in this dataset can be used to align, and crop faces or as ground truth for a face detection algorithm. The images are from different resolutions and settings, example portrait images, groups of people, etc.
- Airplanes: The dataset used for the Airplanes category is The FGVC-Aircraft dataset. The FGVC-Aircraft is a benchmark dataset for categorisation of aircraft and contains about 10200 images of aircraft.

In this paper, 750 images were shuffled and taken from the dataset of each category as inputs for the training set, and 250 images, which were again shuffled from the original dataset, were taken as inputs for the testing set for the conventional Test-Train split model. In the K-fold and Leave One Out validation methods, the same data is re-used according to their respective techniques. In K-fold, the model is trained using K-1 folds of the dataset, and the Kth fold is used to validate the model. In Leave One Out validation, K is equal to the number of samples in the dataset, and each input is used as a test case for the model individually.

*3.3 GPU Implementation*

The GPUs used for the processing of this model are the NVIDIA GeForce 900 series, which are preferred over the older GPUs since they have an R/W L2 global cache for device memory. Due to this, the speed of the programs is greatly boosted, and thus, it simplifies writing the code, without having to worry about speed bottlenecks.

The manual optimization of the CUDA code is very time-consuming. It is often prone to errors; thus, we have optimized it to operate on the newer L2 based cache, which optimizes the running of the model. We have used the following kinds of optimization-pre-computed NumPy arrays, unrolled loops within template kernels and stridden matrices for faster memory accesses and made use of registers wherever possible.

*3.4 Preprocessing of Input*

An image is a 2-D array of numbers between 0-255, but the given image cannot be fed to the model directly in its original form. Since each image might be of different size and composition, the inputs to the CNN model have to be standardized. Since the inputs to the neural network need to be standardized, and thus the images for the model used in this paper are resized to 100x100 to establish a base size for all inputs. Once resized, the images are converted to grayscale. This because grayscale requires lesser memory as opposed to the typical 3-channel RGB images. Once the images are converted to grayscale, the images are converted to NumPy arrays, which stored as [height, width, channel], and are then passed as inputs to the model. NumPy arrays are preferred since they make it easier for the model to read the data and to reduce memory usage, which consequently makes it faster.

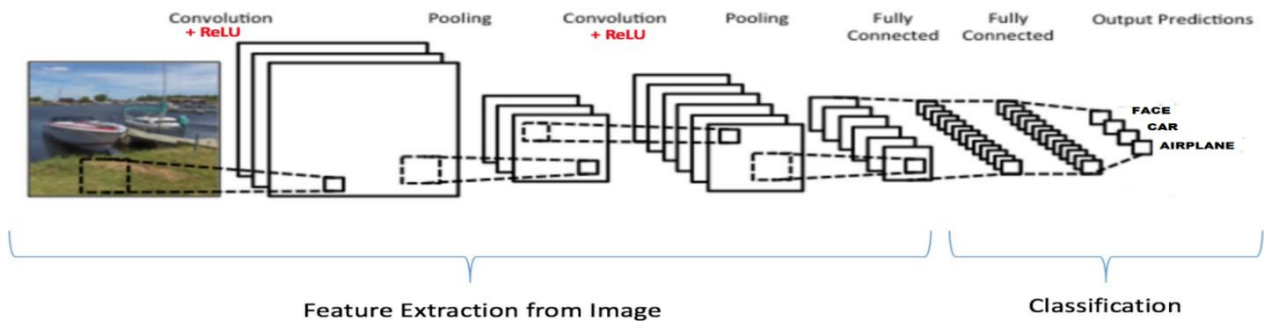

**Fig. 2 –Flowchart of methodology leveraged for model.**

*3.5 Parameters varied*

- Filter size: Input in a neural network is modified with the help of a filter. In CNN's, the filters map moves over individual slices of the image one by one and learns the various parts of the input image. It is a good indicator of how many neighbours' information can be seen whilst processing the input layer. In general, a larger filter size helps produce a model with better performance as compared to a model with a smaller filter size. In this paper, a tabulation of results was performed for various filter sizes such as 5, 7 and 09, and it was observed that the peak performance for the CNN was observed when the filter size was 9.

- Learning Rate: Learning rate can be hypothesized as a hyperparameter which helps control the amount of change in the model which occurs in response to the estimated error which changes each time the weights in the model are updated. Large learning rates can result in undesirable behaviour, while a meagre learning rate would make the training process very slow and would take a very long time to converge. Thus, it is important to find an optimum value for the learning rate, which does not impede learning. In this paper, a tabulation of results was performed for various learning rates such as 0.001, 0.002, 0.004 and 0.008, and it was observed that the peak performance for the CNN was observed when the learning rate was 0.001.

- Number of hidden layers: The layers of neuron between the input and the output layer are called the hidden layers. They take in the weights of the inputs and pass it through the activation function to produce the output. The activation function can be linear or nonlinear. Too many neurons in the hidden layers can lead to over-fitting of the model, and very few hidden layers lead to a poorly trained model which affects the overall accuracy of the CNN Model. Thus, an optimal number of hidden layers must be chosen to exploit the best performance from the model. In this paper, a tabulation of results was performed for a various number of hidden layers such as 2, 3, 4 and 5, and it was observed that the peak performance for the CNN was observed when the number of hidden layers was 2.

- The stride size is the shift (in terms of pixels) in the filter, in one step over the input matrix. The stride size based on the size of the input and the structure of the neural network. The larger the stride size, lesser computation is required as less information is gathered. A larger stride size also leads to a lesser sparse matrix as opposed to having a small stride size. In this paper, a tabulation of results was performed for various stride sizes such as 1, 2 and 3, and it was observed that the peak performance for the CNN was observed when the stride size was 1.

- Epochs: An epoch is when an entire dataset is passed forward and backwards through the neural network once. The number of iterations the neural network goes through in the process of training of the network is called epochs. In this paper, a tabulation of results was performed for various number of epochs such as 5, 10, 15, 20, 25, 30, 35, and 40 and graphs were drawn for the performance of the model with the number of epochs and the accuracy being the parameters plotted on the graph.

- Activation Function (Relu): The activation function of a node defines the output of the node when given an input, or a set of inputs. In this paper, we have used Relu as the activation function for the models. Relu stands for Rectified Linear Unit for a nonlinear operation. The output is $f(x) = \max(0, x)$. The main purpose of Relu is to add non-linearity in the convolutional network since real world data generally requires the network to learn non-negative linear values. Relu stands for Rectified Linear Unit for a nonlinear operation. The output is $f(x) = \max(0, x)$. In terms of performance on the model used in this paper, Relu gives a better performance as compared to other activation functions such as sigmoid and tanh.

*3.6 Cross validation*

Cross-Validation is a very powerful tool since it gives us an idea of how well the model performs on unknown or unseen data. It addition it also helps by putting the data to better use, whilst also providing a greater insight into the performance of the algorithm. The images which are given as inputs to the model are split into training sets and validation sets. The different validation techniques used are discussed below:

- Train-Test Split Method: A typical approach for training a neural network involves splitting the training dataset into three parts, namely train, test, and validation data. This is done employing the train-test split function which is inbuilt in the scikit-learn module in python. In the model used in this paper, the test-train split has been taken as 70-30, where 70 indicates 70 percent of the data has been used for training, and the remaining 30 percent of the data has been used for the purpose of testing.

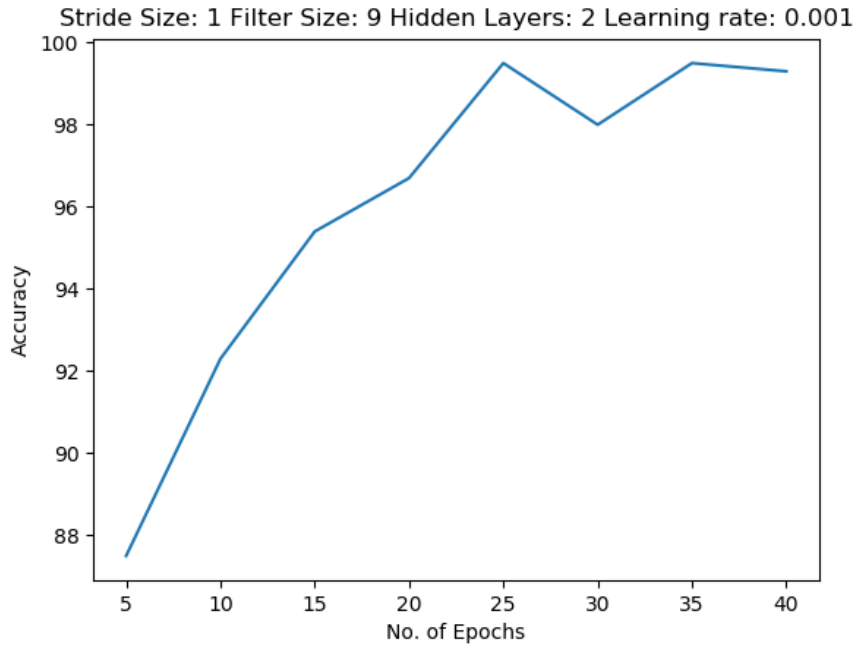

**Fig. 4 –Plot of accuracy vs No of Epochs for Test-Train split method.**

- K-Fold Cross Validation: In this method, the dataset is split into K folds (the value of K is decided based on the size of the dataset). The model is trained using K-1 folds and the Kth fold is used to validate the model. This process is repeated such that all the folds are used as validation (test set). The K-fold technique is used to generate a less biased model, as all the data is tested. For this paper, the value of k has been taken as 5. As k gets larger, the difference in size between the training set and the re-sampling subsets (remaining k-1 batches) gets smaller. As this difference decreases, the bias of the technique becomes smaller.

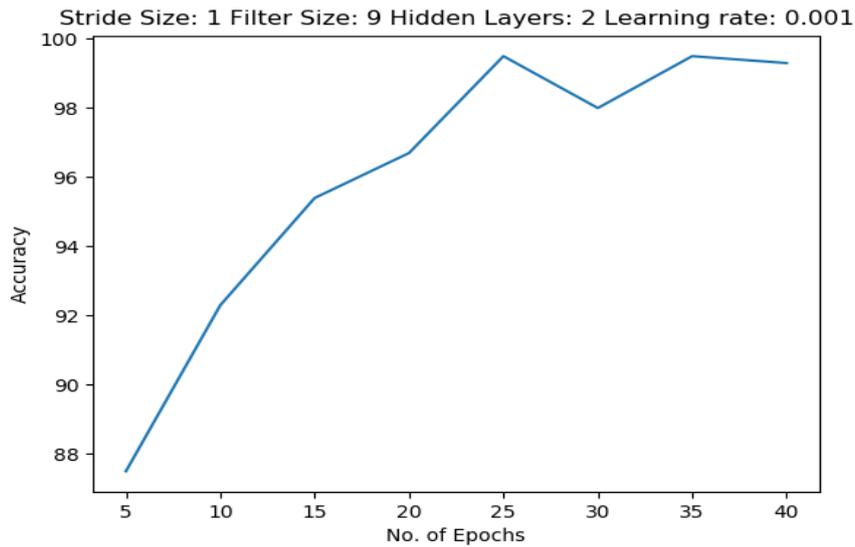

**Fig. 4 –Plot of accuracy vs No of Epochs for K-fold validation method.**

- Leave One Out Validation: This is a special case of cross K-fold validation where K is equal to the number of samples in the dataset. Each input is used as a test case individually. This is more expensive compared to K-Fold Cross Validation but produces a less biased model as compared to K-fold.

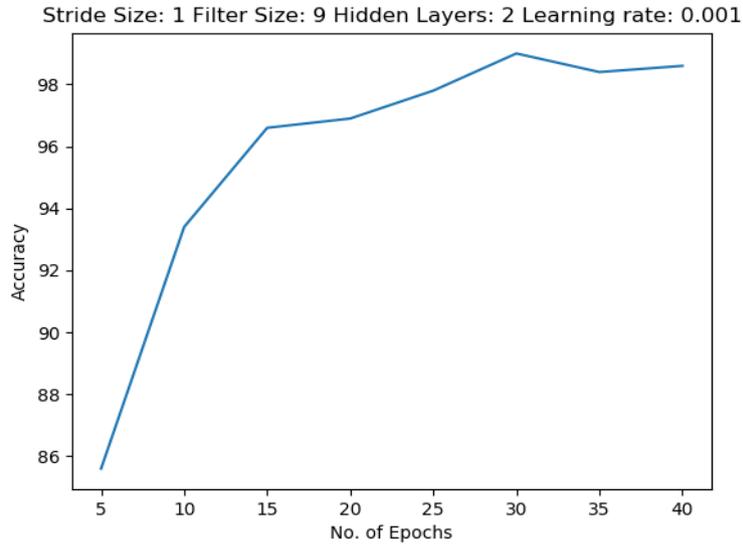

**Fig. 5-Plot of accuracy vs No of Epochs for Leave One Out Validation Method**

## 4. Observations

After finding the optimal parameters for the model, the observations for accuracy vs number of epochs for each of the models are shown below. Through various iterations by trial and error, the optimal parameters for the test train split model, k fold cross validation, and the leave one out validation model are recognized as stride size=1, filter size=9, number of hidden layers=2, and learning rate=0.001. It can be seen that using K-fold and leave one out validation, the accuracy of the model is increased using the same parameters.

**Table 1 – Accuracy obtained with different validation techniques.**

| No. of Epochs | Validation Techniques | | |
|---|---|---|---|
| | **Test-Train Split** | **K-fold** | **Leave One Out** |
| 5 | 60 | 87.5 | 85.6 |
| 10 | 65 | 92.3 | 93.4 |
| 15 | 65.8 | 95.4 | 96.6 |
| 20 | 67.9 | 96.7 | 96.9 |
| 25 | 69.7 | 99.5 | 97.8 |
| 30 | 69.7 | 98 | 99 |
| 35 | 69.5 | 99.5 | 98.4 |
| 40 | 69.8 | 99.3 | 98.6 |

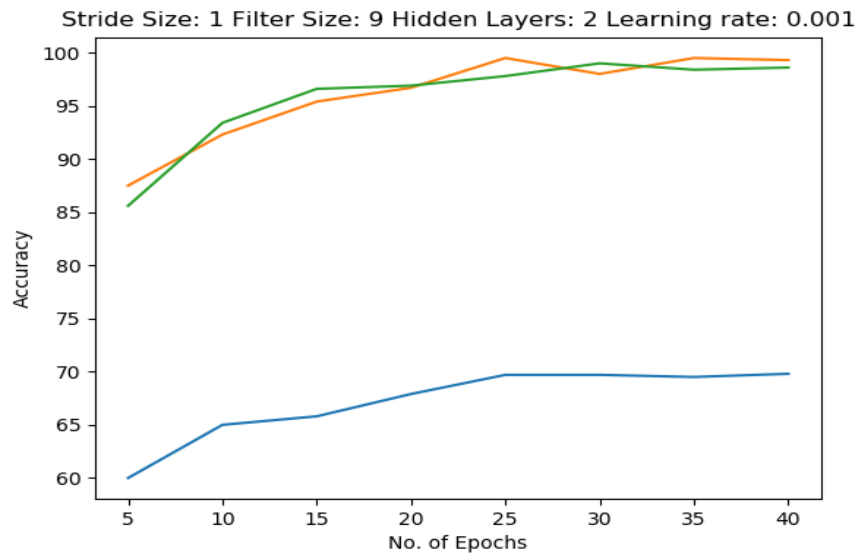

**Fig. 6 –Plot of Accuracy vs No. of Epochs for all models where the blue line indicates the Test-Train split model, the green line indicates the Leave One Out validation graph, and the orange line indicates the K-fold validation model.**

## 5. Conclusion

Through empirical analysis, it is observed that cross-validation techniques such a K-Fold and Leave One Out Validation result in better performances compared to the traditional train-test split. The cross-validation techniques result in an accuracy of 99%, and the traditional train-test model has an accuracy of 68%. When working with a limited dataset, the traditional methods result in biased training as either a small set of data can be used for testing or limited data is used for training. However, the cross-validation techniques make sure that every image of the dataset is subject to prediction at least once while having enough data to train the model. Using cross-validation techniques might lead to overfitting if the model is over-trained and using Leave One Out Validation might lead to higher bias, as the training set is small for each iteration. This paper only focuses on using cross-validation techniques to improve accuracy for object recognition. It can be kept as a base while looking into the novel techniques leveraged for real-world applications.

When compared to the K-fold method, Leave One Out Validation leads to a more unbiased model, but also results in higher computation time as every image is individually set aside for testing. Therefore, Leave One Out Validation is better than K-Fold for smaller datasets. The higher performance of the model, as observed compared to the test-train split method warrants the use of cross-validation techniques.